%% file: acl_latex.tex
\pdfoutput=1

\documentclass[11pt]{article}

\usepackage{url}
\usepackage[T1]{fontenc}

\usepackage{amsmath}
\usepackage{mathtools}
\usepackage{times}
\usepackage{latexsym}
\usepackage{textcomp}
\usepackage{tikz}
\usepackage{pgfplots}
\usepackage{graphbox}
\usepackage{multirow}
\usepackage{longtable}
\usepackage{supertabular,booktabs}
\usepackage{enumitem}
  \usepackage{setspace}
  
\pgfplotsset{compat=1.9}

\definecolor{tikzblue}{RGB}{176,224,230}
\definecolor{tikzgreen}{RGB}{220,251,152}

\definecolor{bblue}{HTML}{4F81BD}
\definecolor{rred}{HTML}{C0504D}
\definecolor{ggreen}{HTML}{9BBB59}
\definecolor{ppurple}{HTML}{9F4C7C}

\usepackage[utf8]{inputenc}

\usepackage{hyperref}
\usepackage{booktabs}
\usepackage{multirow}
\usepackage{footmisc}
\usepackage{microtype}
\usepackage{caption}
\usepackage{subcaption}
\usepackage{tablefootnote}
\usepackage{float}

\input{math-com}

\usepackage[final]{acl}

\usepackage{times,latexsym}

\usepackage{microtype}

\usepackage{inconsolata}

\usepackage{graphicx}

\newcommand{\ours}{FOLIO\xspace} 

\title{\ours: Natural Language Reasoning with First-Order Logic}

\author{
 Simeng Han$^{1}$ 
 \quad \textbf{Hailey Schoelkopf}$^{1}$ 
 \quad \textbf{Yilun Zhao}$^{1}$ 
 \quad \textbf{Zhenting Qi}$^{2}$ \\
 \quad \textbf{Martin Riddell}$^{1}$ 
 \quad \textbf{Wenfei Zhou}$^{3}$ 
  \quad \textbf{James Coady}$^{1}$ 
   \quad \textbf{David Peng}$^{1}$ \\ 
 \quad \textbf{Yujie Qiao}$^{1}$ 
 \quad \textbf{Luke Benson}$^{1}$ 
 \quad \textbf{Lucy Sun}$^{1}$ 
\quad \textbf{Alex Wardle-Solano}$^{1}$ \\
    \quad \textbf{Hannah Szabo}$^{1}$
 \quad \textbf{Ekaterina Zubova}$^{1}$ 
 \quad \textbf{Matthew Burtell}$^{1}$
 \quad \textbf{Jonathan Fan}$^{4}$ \\
 \quad \textbf{Yixin Liu}$^{1}$ 
 \quad \textbf{Brian Wong}$^{1}$ 
 \quad \textbf{Malcolm Sailor}$^{1}$
 \quad \textbf{Ansong Ni}$^{1}$ \\
 \quad \textbf{Linyong Nan}$^{1}$ 
 \quad \textbf{Jungo Kasai}$^{5}$ 
 \quad \textbf{Tao Yu}$^{6}$ 
 \quad \textbf{Rui Zhang}$^{7}$ \\
 \quad \textbf{Alexander R. Fabbri}$^{9}$  
 \quad \textbf{Wojciech Kry\'sci\'nski}$^{9}$ \\
  \quad \textbf{Semih Yavuz}$^{9}$  
  \quad \textbf{Ye Liu}$^{9}$  
 \quad \textbf{Xi Victoria Lin}$^{8}$ 
  \quad \textbf{Shafiq Joty}$^{9}$ 
  \quad \textbf{Yingbo Zhou}$^{9}$ \\
 \quad \textbf{Caiming Xiong}$^{9}$ 
 \quad \textbf{Rex Ying}$^{1}$ 
 \quad \textbf{Arman Cohan}$^{1}$ 
 \quad \textbf{Dragomir Radev}$^{1,9}$ \\
  $^1$Yale University, 
  $^2$Harvard University, 
  $^3$NVIDIA, 
  $^4$Iowa City West High School \\
  $^5$University of Washington,
  $^6$University of Hong Kong \\
  $^7$Penn State University,
  $^8$Meta AI,
  $^9$Salesforce Research 
 }

\include{main/tables/figures}

\begin{document}
\maketitle

\input{main/0-abstract}

\input{main/1-introduction}

\input{main/2-related_work}

\input{main/3-dataset_construction}

\input{main/4_task_definition}

\input{main/5_methods}

\input{main/6-experiments}
\input{main/7-analysis}

\input{main/8-conclusion}
\bibliography{acl}
\appendix
\input{Appendix}

\end{document}

%% file: math-com.tex
\usepackage{amsmath}
\usepackage{amsfonts,bm}
\usepackage{xspace}

\newcommand{\ie}{{\em i.e.,}\xspace}

\definecolor{mypink3}{cmyk}{0, 0.7808, 0.4429, 0.1412}

\makeatletter   
\newcommand{\sveryshortarrow}[1][3pt]{\mathrel{%
    \vcenter{\hbox{\rule[-.5\fontdimen8\scriptfont3]
               {\scriptratio\dimexpr#1\relax}{\fontdimen8\scriptfont3}}}%
   \mkern-4mu\hbox{\let\f@size\sf@size\usefont{U}{lasy}{m}{n}\symbol{41}}}}
\makeatother

\def\eqref#1{equation~\ref{#1}}

\def\1{\bm{1}}

\def\m1{{\bm{1}}}

\DeclareMathAlphabet{\mathsfit}{\encodingdefault}{\sfdefault}{m}{sl}
\SetMathAlphabet{\mathsfit}{bold}{\encodingdefault}{\sfdefault}{bx}{n}



%% file: main/tables/figures.tex
\newcommand{\dalechall}{
\begin{figure}[!t]
    \centering
    \includegraphics[width=\columnwidth]{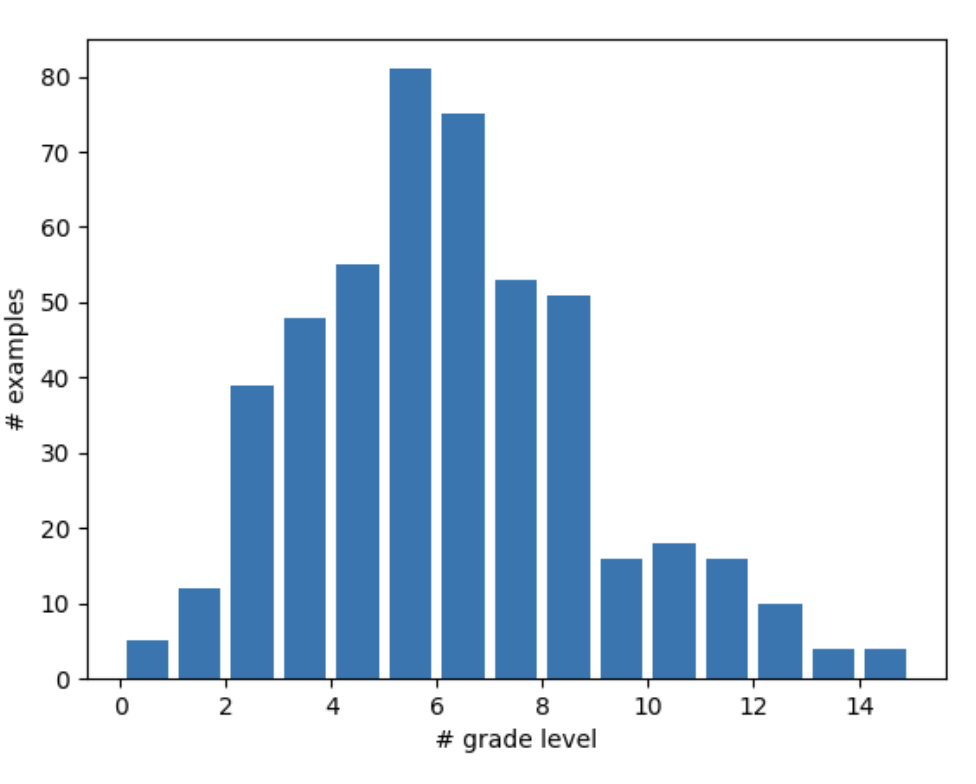}
    \caption{Dale-Chall Readability Distribution.}
    \label{fig:complexity}
\end{figure}
}

\newcommand{\cf}{
\begin{figure}[!t]
    \centering
    \includegraphics[align=t,scale=0.21]{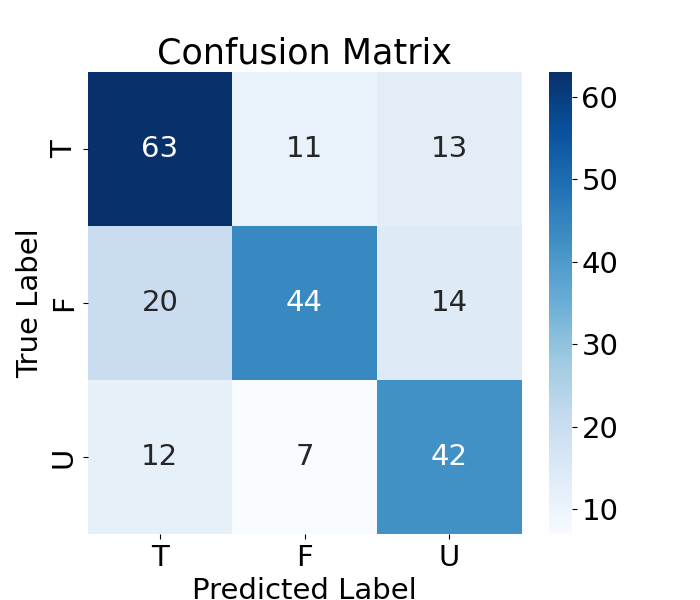}
    \includegraphics[align=t,scale=0.21]{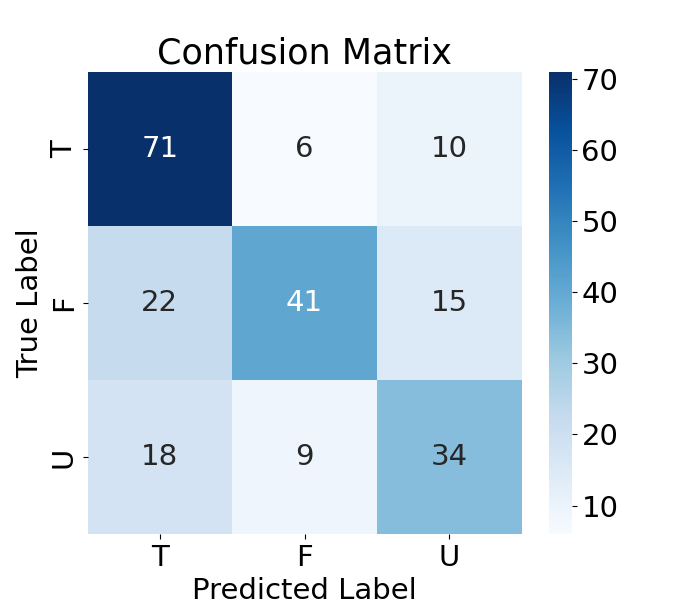}
    \caption{Confusion matrices for the results of fine-tuning RoBERTa-Large and few-shot prompting GPT-4.}
    \label{fig:cf}
\end{figure}
}

\newcommand{\wikihyb}{
\begin{table}[!t]
\centering
\small
\begin{tabular}{llccc}
\toprule
 \textbf{Method} & \textbf{Model} & Wiki & Hyb \\
\midrule
\textit{Fine-tuning} & RoBERTa-large & 60.71 & 63.48 \\
\midrule
\textit{NL Prompting} & GPT-3.5-Turbo & 68.88 & 47.70  \\
& GPT-4 & 75.43 & 53.10  \\
\midrule
\textit{NL-FOL ExcAcc} & GPT-3.5-Turbo & 45.17 & 61.82 \\
& GPT-4 & 59.12 & 67.93 \\
\bottomrule
\end{tabular}
\caption{Performance differences on the WikiLogic and HybLogic subset of \ours. WikiLogic has more diverse logical structures while HybLogic stories have higher reasoning depths.}
\label{tab:wikihyb}
\end{table}
}

\newcommand{\nlandfolprompt}{
\begin{table}[!t]
\centering
\small
\begin{tabular}{lcccc}
\toprule
\textbf{Model} & \textbf{NL} &  \textbf{NL-FOL} & \textbf{FOL}  & \textbf{NL+FOL} \\
\midrule
\textit{GPT-3.5} &  58.34 &  55.96 & 57.92 & 57.75 \\
\textit{GPT-4} & 64.16 & 63.82 & 64.01 & 65.21 \\
\bottomrule
\end{tabular}
\caption{Comparison of the results across different input formats with few-shot prompting. NL, NL-FOL, FOL, NL + FOL stands for NL prompting, execution accuracy of NL-FOL translation, using only FOL in the prompt and using concatenated NL and FOL in the prompt respectively.}
\label{tab:nl+fol-prompt}
\end{table}
}

\newcommand{\mainresult}{
\begin{table}[!t]
\centering
\small
\begin{tabular}{llc}
\toprule
\textbf{Model} & \textbf{Size} & \textbf{Acc (\%)}  \\
\midrule
majority baseline & - & 38.5\% \\
random probability & - & 33.3 \% \\ 
\midrule
\textit{Fully supervised fine-tune}\\
\midrule
BERT-base & 110M & 56.8 \\
BERT-large & 340M & 59.0 \\
RoBERTa-base  & 110M & 56.8\\
RoBERTa-large & 340M & 62.1 \\
Flan-T5-Large & 783M & \textbf{65.9} \\ 
\midrule
\textit{0-shot NL Prompt} \\ 
\midrule
 GPT-3.5-Turbo & - & 53.1  \\
 GPT-4 & - & 61.3 \\ 
\midrule
\textit{8-shot NL Prompt} \\ 
\midrule
 LLama-13B & 13B &  33.6  \\
 \midrule 
LLama-70B & 70B &  44.0 \\
LLama-70B - CoT & 70B & 47.8 \\
LLama-70B - ToT & 70B & 48.4 \\
\midrule
text-davinci-002 & - & 49.5 \\
 GPT-3.5-Turbo & - & 58.3 \\
 GPT-4 & - & 64.2 \\ 
 GPT-4 - CoT \citeyearpar{wei2022chain} & - & 68.9 \\ 
 GPT-4 - CoT with SC \citeyearpar{wang2023selfconsistency} & - & 69.5 \\ 
 GPT-4 ToT \citeyearpar{yao2023tree} & - & 70.0 \\
 \midrule 
 \textit{LR-specific Methods} \\ 
 \midrule
 Logic-LM \citeyearpar{pan-etal-2023-logic}&  - & 78.1\\
 LINC  \citeyearpar{linc} & - & 73.1 \\ 
 DetermLR \citeyearpar{sun2023indeterminacy} & - & 77.5 \\
 
\bottomrule
\end{tabular}
\caption{Logical reasoning results of fully supervised fine-tuning and few-shot prompting on \ours test set. The model sizes of text-davinci-002, GPT-3.5-Turbo and GPT-4 are hidden from public\footnotemark. CoT stands for chain-of-thought prompting \citep{wei2022chain}. SC stands for self-consistency \citep{wang2023selfconsistency}. ToT stands for tree-of-thought prompting \citep{yao2023tree}. 
}
\label{table:main_folio}
\end{table}
}

\newcommand{\nltofol}{
\begin{table}[!t]
\centering
\setlength{\tabcolsep}{3pt}
\small
\begin{tabular}{lcccc}
\toprule
 \multirow{2}{*}{ \space\space\space\space \textbf{Model}} & \multicolumn{2}{c}{\textbf{Zero-Shot} } & \multicolumn{2}{c}{\textbf{Few-Shot} } \\
\cmidrule(lr){2-3} \cmidrule(lr){4-5} 
 & Synv & ExcAcc & Sync & ExcAcc \\
\midrule
GPT-3.5-Turbo & 68.4	& 50.4 & 93.3 & 56.0\\
GPT-4 & 86.1 & 51.7 & 93.9 & 63.8 \\
\bottomrule
\end{tabular}
\caption{NL-FOL translation results on \ours. 
SynV measures syntactic validity and ExcAcc measures the inference engine execution accuracy.}
\label{tab:nl-fol-metrics}
\end{table}

}

\newcommand{\casestudy}{

\begin{table}[t!]
    \centering
    \setlength{\tabcolsep}{2pt}
    \scalebox{0.75}{
    \begin{tabular}{llllll}
    \toprule
\multicolumn{3}{l} {\textbf{NL Premises}} & \multicolumn{3}{l}{\textbf{NL Conclusions}} \\ 
\multicolumn{3}{l} {1. A moth is not a butterfly.} & \multicolumn{3}{l} {A. Cerura vinula emerges} \\ 
\multicolumn{3}{l} {2. Butterflies have thin antennae.} &  \multicolumn{3}{l}{from cocoons.} \\ 
\multicolumn{3}{l} {3. Moths emerge from cocoons.} & \multicolumn{3}{l} {B. Cerura vinula does not} \\
\multicolumn{3}{l} {4. Some moths are pests.} & \multicolumn{3}{l} {have thin antennae.}\\
\multicolumn{3}{l} {5. Cerura vinula is a moth.}  & \multicolumn{3}{l} {C. Cerura vinula is a pest.}\\

\midrule

\multicolumn{1}{l}{\textbf{Labels}} & \multicolumn{1}{l}{\textbf{GPT-4}} & \multicolumn{1}{l}{\textbf{Fine-tune}} \\ 
 
\multicolumn{1}{l}{A. True} &  \multicolumn{1}{l}{True}  & \multicolumn{1}{l}{Unknown} \\ 
 
\multicolumn{1}{l}{B. Unknown} &  \multicolumn{1}{l}{True} & \multicolumn{1}{l}{True} \\ 
 
\multicolumn{1}{l}{C. Unknown} & \multicolumn{1}{l}{Unknown} & \multicolumn{1}{l}{True} \\ 
 
 \bottomrule
    \end{tabular}
    }
    \caption{A WikiLogic story and model predictions. }
    \label{tab:case_study1}
\end{table}
}

\newcommand{\casestudytwo}{
\begin{table}[t!]
    \centering
    \small
    \scalebox{0.75}{
    \begin{tabular}{l}
\toprule
\textbf{NL Premises} \\
1. Some employees good at time management do not exercise \\ every week. \\
2. All employees good at time management are efficient in \\ dealing with daily work. \\
3. All employees efficient in dealing with daily work 
perform \\ better than others. \\
4. All employees who perform better than others have more \\ opportunities to get a promotion. \\
5. James does not have more opportunities to get a promotion.\\
\midrule
\textbf{NL Conclusions} \\
A. James exercises every week. \\
B. James exercises every week and is good at time management. \\
C. If James does not perform better than others, then he \\exercises every week and is good at time management. \\
\midrule

 \textbf{Labels} \space\space\space\space\space\space  \space\space\space\space\space\space\space \space \space \space \space \textbf{GPT-4}\space\space\space\space\space\space  \space\space\space\space\space\space \textbf{Fine-tune} \\ 
 
 A. Unknown \space\space\space\space\space\space\space\space Unknown \space\space\space\space\space\space Unknown \\ 
 
 B. False \space\space\space\space\space\space  \space\space\space\space\space\space\space\space\space\space Unknown \space\space\space\space\space\space False \\ 
 
 C. False \space\space\space\space\space\space  \space\space\space\space\space\space\space\space\space\space True \space\space\space\space\space\space  \space\space\space\space\space\space\space\space True \\ 
 
 \bottomrule
    \end{tabular}
    }
    \caption{A HybLogic story and model predictions.}
    \label{tab:case_study2}
\end{table}
}

%% file: main/0-abstract.tex
\begin{abstract}
Large language models (LLMs) have achieved remarkable performance on a variety of natural language understanding tasks. However, existing benchmarks are inadequate in measuring the complex logical reasoning capabilities of a model. We present \ours, a human-annotated, logically complex and diverse dataset for reasoning in natural language (NL), equipped with first-order logic (FOL) annotations. 
\ours consists of 1,430 examples (unique conclusions), each paired with one of 487 sets of premises used to deductively reason for the validity of each conclusion.
The logical correctness of the premises and conclusions is ensured by their FOL annotations, which are automatically verified by an FOL inference engine.
In addition to the main NL reasoning task, NL-FOL pairs in \ours constitute a new NL-FOL translation dataset.
Our experiments on \ours systematically evaluate the FOL reasoning ability of supervised fine-tuning on medium-sized language models. 
For both NL reasoning and NL-FOL translation, we benchmark multiple state-of-the-art language models. Our results show that a subset of \ours presents a challenge for one of the most capable {Large Language Model (LLM)} publicly available, GPT-4. 
\end{abstract}

%% file: main/1-introduction.tex
\section{Introduction}

\begin{table*}[t!]
\resizebox{\textwidth}{!}{%
\begin{tabular}{lrllccccc}
\toprule
\textbf{Dataset} &
  \multicolumn{1}{l}{\textbf{Size}} &
  \multicolumn{1}{l}{\textbf{Reasoning}} &
  \multicolumn{1}{c}{\textbf{Text Source}} &
  \multicolumn{1}{c}{\textbf{\begin{tabular}[c]{@{}c@{}}Real-World \\ Resources\end{tabular}}} &
  \multicolumn{1}{c}{\textbf{\begin{tabular}[c]{@{}c@{}} \# Reasoning \\ Depth \end{tabular}}} & 
  \multicolumn{1}{c}{\textbf{\begin{tabular}[c]{@{}c@{}}Vocab \end{tabular}}} & 
  \multicolumn{1}{c}{\textbf{\begin{tabular}[c]{@{}c@{}} \# Distinct \\ AST \end{tabular}}}
  \\
\midrule
CLUTTER \citeyearpar{sinha-etal-2019-clutrr}  & 6k                  & Inductive   & Synthetic        & \texttimes & \texttimes   & - &   \texttimes      \\
RECLOR \citeyearpar{Yu2020ReClor}   & 6k    & Mixed forms   & GMAT, LSAT exams & $\checkmark$ & \texttimes    & - &   \texttimes     \\
LogiQA \citeyearpar{liu2021logiqa} & 8.6k   & Mixed forms   & NCSE exams       & $\checkmark$ & \texttimes    & - &   \texttimes        \\
RuleTaker \citeyearpar{ruletaker}  & 500k       &  Deductive   & Synthetic        & \texttimes & $0\sim5$ & 101 & 48      \\
ProofWriter \citeyearpar{tafjord-etal-2021-proofwriter} & 500k         & Deductive  & Synthetic        & \texttimes & $0\sim5$ & 101 & 48      \\
LogicNLI \citeyearpar{tian-etal-2021-diagnosing}  & 20k     & FOL & Synthetic        & \texttimes & $1\sim5$  & 1077   &  30 \\ 
BigBench \citeyearpar{srivastava2022beyond}  &  1300    & Mixed forms & Human-Written  & Partially & \texttimes &  - & - \\ 
ProntoQA \citeyearpar{saparov2023language}  &  200   & Deductive &  Synthetic & $\checkmark$ & 1, 3, 5 & - & - \\ 

\midrule
\textbf{\ours (ours)} &
  \multicolumn{1}{r}{1,435} &
  \multicolumn{1}{l}{FOL} &
  \multicolumn{1}{c}{\textbf{Expert-written}} &
  \multicolumn{1}{c}{$\checkmark$} &
  \multicolumn{1}{c}{\boldsymbol{$0\sim7$}} &  \textbf{4351} & \textbf{76} \\
\bottomrule
\end{tabular}
}
\caption{Comparison of \ours with other datasets related to logical reasoning. 
\#Distinct AST stands for the number of distinct abstract syntax trees, representing the number of distinct sentence-level logic structures in the corpus. \ours is the first expert-written dataset for FOL reasoning equipped with parallel FOL formulas. The examples are mostly aligned with real-world knowledge and use highly natural wordings. It also has a greater variety than the previous datasets in terms of reasoning depths with a larger number of distinct logic patterns and a large vocabulary. 
}
\label{table:comparison}
\end{table*}

\begin{table*}[t!]
\centering
\resizebox{\textwidth}{!}
{%
\begin{tabular}{ll}
\toprule
A \ours example based on the Wild Turkey Wikipedia page: ~\url{https://en.wikipedia.org/wiki/Wild_turkey} & \\
\midrule
\textbf{NL premises} & \textbf{NL Conclusions -> Labels} \\
1. There are six types of wild turkeys: Eastern wild turkey, Osceola wild turkey, Gould’s wild turkey,  & A. Tom is an Ocellated wild turkey. -> True \\
Merriam’s wild turkey, Rio Grande wild turkey, and the Ocellated wild turkey. & B. Tom is an Eastern wild turkey. -> False \\
2. Tom is not an Eastern wild turkey. &  C. Joey is a wild turkey. -> Unknown \\
3. Tom is not an Osceola wild turkey. &   \\
4. Tom is also not a Gould's wild turkey. \\ 
5. Tom is neither a Merriam's wild turkey, nor a Rio Grande wild turkey. & \\ 
6. Tom is a wild turkey. & \\

\midrule

\textbf{FOL Premises}  & \textbf{FOL conclusions -> Labels} \\

1. $\forall x (\text{WildTurkey}(x) \rightarrow (\text{EasternWildTurkey}(x) \lor \ \text{OsceolaWildTurkey}(x) \lor \ \text{GouldsWildTurkey}(x)$ & A. $\text{OcellatedWildTurkey}(tom)$ -> True \\

$\lor \ \text{MerriamsWildTurkey}(x) \lor \ \text{RiograndeWildTurkey}(x) \lor \ \text{OcellatedWildTurkey}(x)))$ &  B. $\text{EasternWildTurkey}(tom)$ -> False \\ 

2. $\lnot \text{EasternWildTurkey}(tom)$ & C. $\text{WildTurkey}(joey)$ -> Unknown \\ 

3. $\lnot \text{OsceolaWildTurkey}(tom))$ &  \\
 
4. $\lnot\text{GouldsWildTurkey}(tom) $ & \\ 
5. $\lnot \text{MerriamsWildTurkey}(tom) \land \lnot \text{RiograndeWildTurkey}(tom)$ \\ 
6. $\text{WildTurkey}(tom)$ & \\

\bottomrule
\end{tabular}
}
\captionof{table}{An example story in \ours based on the knowledge from the Wikipedia page on wild turkeys. 
The story consists of five premises and three conclusions with their corresponding FOL formulas and labels for the conclusions. 
All five premises are needed to infer the conclusions. The model needs to reason under logic patterns with universal quantification ($\forall$), negation ($\lnot$), conjunction ($\land$), and disjunction ($\lor$).
}
\label{tab:example}
\end{table*}

Large language models (LLMs) have achieved remarkable performance on a variety of natural language tasks \citep{openai2023gpt4, touvron2023llama, srivastava2023imitation, superglue}. Logical reasoning is a central component for intelligent systems and should be sufficiently and independently evaluated \citep{russell-norvig-2010}. However, existing natural language tasks are inadequate in measuring the complex logical reasoning capability of a model \citep{ srivastava2023imitation, saparov2023language, tian-etal-2021-diagnosing}.

Several datasets related to logical reasoning have recently been proposed. 
However, existing benchmarks often exhibit limited complexity in reasoning or lack language naturalness. 
Some of these common benchmarks do not specifically evaluate logical reasoning independently of other forms of reasoning \citep{Yu2020ReClor, liu2021logiqa}. Those specifically designed for measuring logical reasoning are insufficient in terms of logical reasoning complexity and natural language variety. As shown in Table~\ref{table:comparison}, examples in RuleTaker \citep{ruletaker} and LogicNLI \citep{tian-etal-2021-diagnosing} need at most five depths of reasoning. The entire corpus of RuleTaker or LogicNLI has fewer than 50 distinct abstract syntax trees. RuleTaker has only 101 words in its vocabulary and LogicNLI has 1077 words in the vocabulary. Moreover, none of them are written by humans with information drawn from real-world knowledge, making them less applicable to real-world reasoning scenarios. The logical deduction portion in BigBench \citep{srivastava2023imitation} requires commonsense reasoning besides logical deduction. ProntoQA \citep{saparov2023language} only contains logical reasoning questions that are answerable with repeated applications of the Modus Ponens inference rule. 

We present a natural language reasoning dataset, \emph{\ours}, with first-order logic reasoning problems which require the models to decide the correctness of conclusions given a {\it world} defined by the premises. 
In \ours, we aim to ensure high language naturalness and complexity, an abundant vocabulary, and factuality while also maintaining high reasoning complexity. \ours is a high-quality and manually curated dataset, written by CS undergraduate and graduate students and researchers in academia and industry. 
To ensure the conclusions of our examples follow the premises logically, we annotated all reasoning examples with first-order logic (FOL) formulas. An example of \ours is shown in Table~\ref{tab:example}. Based on our annotations, we propose a new NL-FOL translation task where an NL reasoning example is translated into its FOL counterpart.
Finally, we benchmark the performance of strong LMs in both fully supervised and few-shot settings to understand their capabilities in logical reasoning {(\ie\ deriving the truth value of a logical conclusion from NL premises)}. 
Under the few-shot setting, the most capable publicly available LLM so far achieves only 53.1\% on the stories written in a hybrid manner, which is slightly better than random. 

To sum up, the contributions of this paper are threefold.
1) We release a natural language reasoning dataset written by expert annotators, \ours, with first-order logical reasoning problems.
2) We use formal logic, i.e., FOL to ensure the logical validity of the examples written in NL and propose a new NL-FOL translation task. 
3) We benchmark the performance of LMs by fine-tuning models and prompting LLMs with few-shot examples, on the \ours reasoning task.
We hope that \ours, as a challenging logical reasoning dataset, will be used to facilitate measuring progress in the logical reasoning capabilities of language models.


%% file: main/2-related_work.tex
\section{Related Work}
\subsection{Datasets for reasoning from text}
Developing models that can reason in texts has been a core goal in NLP since the field’s early days~\cite{cooper1996using}. Since then, there has been massive progress in reasoning over text. Various benchmarks that focus on different aspects of reasoning over textual inputs are proposed, including natural language inference (NLI)~\cite{bowman-etal-2015-large, wang2019superglue}, reasoning for commonsense knowledge~\cite{talmor-etal-2019-commonsenseqa, he-etal-2021-winologic} and multi-hop reasoning~\cite{yang-etal-2018-hotpotqa, chen-etal-2020-hybridqa}. 
Among these reasoning abilities, logical reasoning has recently attracted an increasing amount of study. ReClor~\cite{Yu2020ReClor} and LogiQA~\cite{liu2021logiqa} both collected multiple-choice questions from standardized graduate admission examinations, answering which requires various types of logical reasoning. However, these datasets cover mixed forms of reasoning and are not intended to test logical reasoning in isolation.

Meanwhile, testing logical reasoning in isolation without involving other forms of reasoning has also attracted researchers in recent years. CLUTRR~\cite{sinha-etal-2019-clutrr} covers inductive reasoning, which is beyond the scope of first-order logic. Synthetic corpuses of deductive reasoning are proposed to evaluate the deductive reasoning ability of pretrained LMs~\cite{clark2021transformers, saeed-etal-2021-rulebert, tian-etal-2021-diagnosing}. However, these datasets do not contain highly natural sentences and often cover limited forms of logic while FOL is much more expressive. \citet{kazemi2023boardgameqa} created a dataset for reasoning with contradictory information. \citet{kawabata2023evaluating} crowdsourced rationales for over 3000 examples based on ReClor \citep{Yu2020ReClor}. 
ProntoQA \citep{saparov2023language} is comprised solely of logical reasoning queries that can be resolved through applying the Modus Ponens inference rule while FOLIO questions require applications of multiple types of inference rules. As shown in Table \ref{table:comparison}, \ours is the first large-scale first-order logic (FOL) reasoning dataset with formal logic annotations in FOL. \ours is logically diverse and complex with complex natural language sentences and a rich vocabulary. 


\subsection{Reasoning using large language models}

Reasoning has been demonstrated as one of the emergent abilities of LLMs of sufficient scale recently \cite{talmor2020leap,  wei2022emergent, chowdhery2022palm}. One such emergent behavior, Chain-of-Thought prompting \cite{wei2022chain}, consists of a series of intermediate reasoning steps output by an LLM. This improves the performance on arithmetic, commonsense, and symbolic reasoning benchmarks significantly. There has been a line of research continuing on from Chain-of-Thought ~\cite{kojima2022large, li2022advance, yao2023tree} to elicit reasoning behavior from LLMs. 
Building on Chain-of-Thought prompting, many techniques used on top of LLMs to improve downstream performance have been formalized into control flows and programs. These are called language model cascades \cite{cascades}, subsuming techniques such as Chain-of-Thought prompting, STaR \cite{star}, and Selection-Inference \cite{creswell2022selection} for reasoning.
~\citet{dasgupta2022language} studied the reasoning ability of LLMs but only used a small set of 48 syllogisms with only two premises each.  \citet{saparov2023language} created a synthetic dataset that and showed that LLMs are capable of making correct individual deduction steps.

With \ours, we aim to set a high standard, ensuring that achieving high performance through superficial strategies and shallow heuristics is prevented, allowing a robust evaluation of the first-order logic reasoning capabilities of LLMs. 
We show that many LLMs fall short on complex first-order logic reasoning, and that significant room for improvement in this area remains.

%% file: main/3-dataset_construction.tex
\section{\ours Corpus Construction}

\label{subsec:annotation}
We collected \ours through a carefully designed manual annotation process to achieve high-quality examples that necessitate complex logical reasoning. Writing natural language reasoning stories with FOL requires sufficient knowledge in both semantic parsing and first-order logic, as well as strong analytical skills. Given the complexities of such annotations, we selected annotators based on a few important criteria to ensures that our dataset is annotated with the highest level of precision and expertise, reflecting the complexity and nuance required for first-order logical reasoning. 1). Our annotators are either college or graduate students who are native English speakers or possess near-native proficiency in English.4 2). They possess formal education in first-order logic, having either completed relevant coursework or undertaken self-directed studies in first-order logic or semantic parsing. At the NL quality check stage, only annotators who are experts in natural language processing or computational linguistics are involved. For the FOL quality check, only annotators who are experts in first-order logic are involved. We also give the annotators several training sessions on how to write a story, by providing them with detailed annotation guidelines. All stories and FOL annotations in FOLIO are written and reviewed by expert annotators, including CS undergraduate and graduate students, and senior researchers, who met the aforementioned criteria. 

We develop our dataset in six stages: WikiLogic collection, HybLogic collection, NL quality control, FOL quality control,  NL-FOL alignment and FOL verification, spending 980 man-hours in total. 


\subsection{Example collection}
\label{sec:example_collection}
We collected our dataset using two different methods in order to obtain examples that are both logically diverse and complex and have abundant abstract syntax tree (AST) variations. The annotators are free to write stories based on any topic they want while writing the stories. 

\paragraph{WikiLogic: annotation from scratch using Wikipedia articles as seeds.}
At this annotation stage, the annotators are asked to select random Wikipedia pages by repeatedly using the Wikipedia Special Random link.\footnote{\url{https://en.wikipedia.org/wiki/Special:Random}} The Wikipedia articles are used to develop ideas for topics to write new stories. We ask the annotators to create new stories \textit{from scratch without using templates} based on real-world knowledge, which should be plausible in general. Each of the stories is composed of several premises and conclusions with truth values of True, False, or Unknown (see Table \ref{tab:example} for an example). We also ask the annotators to write parallel FOL sentences for both the premises and conclusions. 
This results in a wide range of topics, abundant AST variations, and a wide vocabulary for \ours. Table~\ref{table:comparison} shows a comparison of \ours with other reasoning datasets that purely evaluate first-order logic or deductive reasoning. 

\paragraph{HybLogic: hybrid annotation}
The task of generating logically sound stories from scratch for a set of facts is very time-consuming for human writers, where the main challenge is to create complex and varied logical patterns to arrive at a conclusion. To address the problems of solely using manual annotation, we also consider a hybrid approach to facilitate the process. Our hybrid method is based on a common form of logical stories: \textit{syllogisms}. A syllogism consists of two premises and a single conclusion, and the conclusion states some facts about the entities and categories in the premises. 

In this approach, we first generate logically valid stories, which are templates containing abstract categories and entities, by combining multiple syllogisms into a single story template: the conclusion of one syllogism is used as a premise for the next syllogism.
There are 256 logically distinct types of syllogisms and 24 of them are valid \citep{annelehman}. 
We use various combinations of 24 valid syllogisms. We also add in conjunction, disjunction, and implication. We show an example of the resulting templates in Appendix \ref{app:template}. We then ask human annotators to assign nouns, phrases, or clauses to the abstract entities or categories that reflect real-life scenarios to each template and write logically-valid stories in natural language. The usage of the template is to ensure that we have a set of varied and complex logical stories with multiple conclusions.
There are many ways of expressing the same logic template in natural language, and so the generated templates augment, rather than limit, the creativity of humans. 

\subsection{Quality control for NL sentences}
To ensure the highest quality of the dataset, we dedicated considerable attention to the following key aspects of the natural language sentences during the quality control process. 

\paragraph{Factuality and bias}
Our dataset prioritizes realism and factual accuracy, steering clear of biases and stereotypes linked to identity markers like race, ethnicity, gender, sexuality, nationality, class, and religion. Toward these objectives, we manually screened all stories and found that 39.2\% of the stories suffer from at least one of these issues. We implemented a detailed protocol to rewrite these stories. The protocol is in Appendix~\ref{app:factualityprotocol}.


\paragraph{Language quality}
Apart from grammar, we make sure the sentences in our dataset are highly natural. All the sentences are first checked with a grammar checking tool, Grammarly. Our annotators who have graduated from or are senior students studying English Literature conducted a thorough round of review for grammatical correctness and language naturalness. We also eliminate natural language ambiguity when it is possible.  We include rules on eliminating ambiguity in Appendix~\ref{app:languagequality}. Employing these rules effectively reduces the ambiguity of natural language in this reasoning dataset, but incurs the tradeoff of limiting variations in some usage of language. However, we note that there is still sufficient variation in terms of sentence structures and logical structures as shown in Table~\ref{table:comparison}. 

\subsection{Quality control for FOL formulas}
We adopt the FOL definitions and syntax most widely used in the AI community \citep{russell-norvig-2010}. We include more details on the definition of FOL we consider and the FOL modelling convention in Appendix~\ref{app:firstorderlogic} In preliminary investigations, we found that the human-written FOL formulas suffer from FOL consistency issues, which necessitates an additional round of quality control for FOL formulas.

\paragraph{FOL consistency}

One NL sentence can be translated into FOL through multiple non-equivalent ways. For example, sometimes additional information inferred from a sentence can be represented in FOL, leading to multiple representations.
We therefore design an annotation protocol for FOL translation in order to ensure that our FOL translations are as consistent as possible across all examples in our dataset. We highlight a few important strategies used in the annotation protocol in Appendix~\ref{app:folprotocol}. 

\color{black}

\subsection{NL-FOL alignment review}
\label{sec:NL_FOL_Align_review}

Apart from checking whether NL and FOL express equivalent meanings, we also add necessary commonsense knowledge in both the NL and FOL premises. 
Sometimes humans do not write certain commonsense knowledge in the premises that is required in the FOL reasoning process, which is based solely on the premises given. We add such knowledge as additional premises at this stage. In particular, intrinsic properties of some predicates are required in the FOL reasoning process. For example, "\texttt{LocatedIn(x,y)}" should be transitive and "\texttt{BeFamily(x,y)}" should be symmetric.

\subsection{FOL verification}
\label{sec:FOL_verification}
Recognizing that the FOL formula annotations can be error-prone, we verify the syntactic validity and label consistency of FOL formula annotations with an FOL inference engine. We include the details of the FOL inference engine in Appendix~\ref{app:inferencengine}. 

\subsection{Dataset statistics}\label{sec:stats}


\begin{table*}[t!]
\centering
\resizebox{\textwidth}{!}
{%
\begin{tabular}{@{}lccccccccc@{}}
\toprule
\multirow{2}{*}{\textbf{Source}}  & \multirow{2}{*}{\textbf{\#Stories}} & \multirow{2}{*}{\textbf{\#Premises}} & \multirow{2}{*}{\textbf{\#Conclusions}} & \multicolumn{3}{c}{\textbf{NL} } & \multicolumn{2}{c}{\textbf{Logic} } \\
\cmidrule(lr){5-7} \cmidrule(lr){8-9} 
  &  &  & & Vocab  & \#Words & Complexity  & \#Depth & AST \\ \midrule
WikiLogic & 304       & 1353          & 753 & 3250  & $8.50$   & 0 - 14 grade & 1 - 5 &  51 & \\
HybLogic      &    183         &1054                      &682 & 1902 &  $11.52$ & 0 - 14 grade &  5 - 8 & 25  &  \\ \midrule
Total & 487 & 2407 & 1435 & 4351 & $9.86$ & 0 - 14 grade & 76 & 5-8 \\
\bottomrule
\end{tabular}
}
\caption{Statistics based on different data collection methods of \ours. \emph{\#Words} is the average number of words per NL sentence. 
}
\label{tab:data_stats}
\end{table*}

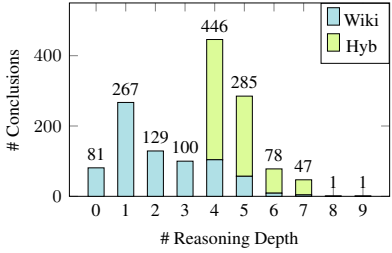
\begin{figure}[!t]
    \centering
    \input{main/figures/nopremise-stacked.tikz}
    \caption{Distribution of reasoning depths}
    \label{fig:nopremise}
\end{figure}

We show basic statistics of \ours and demonstrate the abundant vocabulary and logical complexity of \ours: Tables~\ref{table:comparison}, \ref{tab:data_stats} and Figure~\ref{fig:nopremise}.

\paragraph{Basic statistics}
Table~\ref{tab:data_stats} shows that examples based on Wikipedia make up the largest portion of \ours, with 304 stories, 1,353 NL and FOL premise pairs, and 753 NL and FOL conclusion pairs. Hybrid annotations consist of 183 stories with 1,054 NL and FOL premise pairs, and 682 NL and FOL conclusion pairs in total. 



\paragraph{Natural language complexity}

We use the Dale-Chall Readability Formula \citep{dalechall1, dale1995readability} to show the text complexity of \ours following \citep{singh-etal-2023-misery, arps-etal-2022-hhuplexity, wei-etal-2021-linguistic}. 
We show the distribution of readability in Appendix \ref{app:readability}. 
\paragraph{Logical complexity and diversity statistics}
As shown in Figure~\ref{fig:nopremise}, the mode of reasoning depths is four in \ours. 28.7\% of the examples need five or more depths of reasoning to infer the conclusions, while the previous datasets needed at most five reasoning depths as shown in Table~\ref{table:comparison}.
This illustrates the logical complexity of \ours.
Table~\ref{table:comparison} shows that \ours also has a much larger number of distinct ASTs than the previous datasets, indicating that \ours is much more logically diverse. Figure~\ref{fig:nopremise} demonstrates the distribution of the number of examples in the WikiLogic and HybLogic sets versus the number of premises needed to arrive at a conclusion, showing that most of the conclusions from WikiLogic require one to five premises while those from HybLogic require five to eight premises. 
\paragraph{Vocabulary and topics}
Table~\ref{tab:data_stats} shows that our dataset has a vocabulary of 4,351 words, and the examples based on Wikipedia account for 74\% of the total vocabulary even though the WikiLogic stories take up only 63\% of the total number of stories. The vocabulary of \ours is also significantly larger than the previous synthetically constructed datasets for logical reasoning. 





%% file: main/figures/nopremise-stacked.tikz
\begin{tikzpicture}[scale=0.75]
\begin{axis}[
    width  = 0.45*\textwidth,
    height = 5cm,
    ybar stacked,
    bar width=8pt,
    ymin=0,
    ymax=550,
    legend style={
        font=\small,
        at={(1,1)},
        anchor=north east
    },
    x tick label style={font=\small},
    y tick label style={font=\small},
    symbolic x coords={0, 1, 2, 3, 4, 5, 6, 7, 8, 9},
    ytick style={font=\small},
    xtick=data,
    xlabel style={font=\small},
    ylabel style={font=\small},
    xlabel={\# Reasoning Depth},
    ylabel={\# Conclusions},
    ]
    \addplot[ybar, fill=tikzblue] plot coordinates {(0, 81) (1, 267) (2, 129) (3, 100) (4, 104) (5, 57) (6, 9) (7, 4) (8, 1) (9, 1)};
    \addplot[ybar, fill=tikzgreen] plot coordinates {(0, 0) (1, 0) (2, 0) (3, 0) (4, 342) (5, 228) (6,69) (7, 43) (8,0) (9,0)};
    
    \node[above, font=\small] at (axis cs:0,81) {81};
    \node[above, font=\small] at (axis cs:1,267) {267};
    \node[above, font=\small] at (axis cs:2,129) {129};
    \node[above, font=\small] at (axis cs:3,100) {100};
    \node[above, font=\small] at (axis cs:4,446) {446};
    \node[above, font=\small] at (axis cs:5,285) {285};
    \node[above, font=\small] at (axis cs:6,78) {78};
    \node[above, font=\small] at (axis cs:7,47) {47};
    \node[above, font=\small] at (axis cs:8,1) {1};
    \node[above, font=\small] at (axis cs:9,1) {1};
    
    legend cell align=left,
    \legend{\strut Wiki,\strut  Hyb}
  \end{axis}
\end{tikzpicture}

%% file: main/4_task_definition.tex
\section{Task Definition}
\label{sec:tasks}
We define two new tasks based on \ours, natural language reasoning with first-order logic and NL-FOL translation. 
\subsection{Natural language reasoning with first-order logic}
Each natural language (NL) story $S$ in \ours consists of $n$ premises: $P = \{p_1, p_2,..., p_n\}$ and $m$ conclusions: $H = \{h_1, h_2,..., h_m\}$. All NL stories are annotated with parallel FOL stories $SF$, which are sets of FOL formulas consisting of $n$ premises $PF = \{pf_1, pf_2,..., pf_n\}$ and $m$ conclusions $HF = \{hf_1, hf_2,..., hf_m\}$. $pf_i$ and $hf_i$ are logically and semantically similar to $p_i$ and $h_i$, respectively. {Given $P$ and $H$,} the goal is to determine the truth values of the conclusions: "True", "False" or "Unknown", based on FOL reasoning. 

\subsection{NL-FOL translation}

We propose a new natural language to first-order logic translation (NL-FOL translation) task alongside our reasoning dataset.
The goal of this task is to translate an NL story $S$ to an FOL story $FS$. In particular, each of the NL sentence $p_i$ or $h_i$ and the parallel FOL formula $pf_i$ or $hf_i$ should be logically and semantically equivalent. Moreover, the truth values for the conclusions should be the same based on the NL story $S$ and the parallel FOL story $FS$. In our dataset, the premises and conclusions are set up in such a way to ensure that the inference engine always returns an answer given enough resources such as time and memory. Unlike previous work~\cite{singh2020exploring} which translates problems with a single premise and a single hypothesis, our task is on translating examples of various lengths with a focus on stories with multiple premises. Thus, it also requires the models to consider {\it discourse-level} consistencies as opposed to translation at the sentence level.

\paragraph{NL-FOL evaluation metrics}\label{sec:nl_fol_metric}
Two metrics are adopted to evaluate NL-FOL translation to capture different aspects of the generation results: 1). \textbf{Syntactic validity (SynV)}. The Syntactic Validity score measures whether the FOL formulas are syntactically valid. The score will be 1 if all FOL formulas of an example can pass the syntactic check and 0 otherwise 2). \textbf{Inference Engine execution accuracy (ExcAcc)}. The group of translated FOL for premises and conclusions in one story is fed into our inference engine to output the truth value for each conclusion. 
We define the accuracy of the output labels as the execution accuracy. We leave for future work the design of a more reliable metric of NL-FOL translation.




%% file: main/5_methods.tex
\section{Experiments}
In this section, we describe our experiments and main results.

\subsection{Experimental setup}

\paragraph{Tasks} We conduct experiments on the two tasks in \S\ref{sec:tasks}: \textit{NL reasoning with first-order logic (logical reasoning)} and \textit{NL-FOL translation (NL-FOL).}

\paragraph{Dataset split}
We split \ours by 70\%/15\%/15\% split for the train/validation/test sets with 1,001/203/226 examples respectively. We split by story so that models are evaluated on unseen stories. 

\paragraph{Evaluation metrics}
We use accuracy for evaluating logical reasoning performance. For NL-FOL translation, we use the metrics in Section~\ref{sec:nl_fol_metric}.

\subsection{Models}
We test the logical reasoning capabilities of LMs using fully supervised fine-tuning and few-shot prompting.
We also test NL-FOL translation with few-shot prompting.
\paragraph{Fully supervised fine-tuning} As fine-tuning baselines, we experiment with BERT~\cite{devlin-etal-2019-bert}, and RoBERTa~\cite{liu2020roberta}. We fine-tune the base and large versions of both BERT and RoBERTa, with an additional two-layer classification layer to predict the truth values. For the second task, i.e., NL-FOL translation, we only report few-shot prompting methods.

\paragraph{Few-shot prompting} We conduct zero-shot and few-shot prompting experiments on larger LMs with few-shot capabilities.
For open-source models, we test LLaMA-13B and LLaMA-70B ~\cite{touvron2023llama}, GPT-NeoX-20B~\cite{neox}; for proprietary models we test GPT-3~\cite{gpt3}, GPT-3.5-Turbo and GPT-4 \cite{openai2023gpt4} using prompts with 8 examples.\footnote{In experimenting with different prompts, we found 8 shot examples to perform slightly better. It is also the maximum number of examples that fits in the text-davinci-002 context.}
\paragraph{Prompting strategies} We experiment with incorporating recent prompting strategies into GPT-4 as they have shown improvements in the general reasoning performance of LLMs. The prompting strategies include chain-of-thought (CoT) prompting \cite{wei2022chain}, chain-of-thought prompting with self-consistency \citep{wang2023selfconsistency} and tree-of-thought prompting \citep{yao2023tree}. 

\paragraph{Logical reasoning methods} We also test recent methods specifically designed for logical reasoning: Logic-LM \citeyearpar{pan-etal-2023-logic}, LINC \citep{linc} and DetermLR\citep{sun2023indeterminacy}, using GPT-4 as the base model. 
For the second task (NL-FOL translation), we use the same examples as in the Few-Shot NL experiments except that all the conclusions are included in each example.

We run experiments on five randomly sampled sets of examples from the training set and report the average accuracy.



%% file: main/6-experiments.tex

\subsection{Main results}
\mainresult
\paragraph{Logical reasoning}
The majority baseline of our dataset is 38.5\% since in our test set, there are 87, 78 and 61 examples with labels of true, false and unknown respectively. 
As shown in Table~\ref{table:main_folio}, BERT-base and RoBERTa-base have similar performance on \ours with 56.83\% accuracy. BERT-large has a 2.2\% improvement over BERT-base. RoBERTa-large improves 3.1\% over BERT-large. Flan-T5-Large achieves the highest performance in the fine-tuning setting and the accuracy is 65.7\%. 

We show that zero-shot prompting GPT-3.5 achieves better results than few-shot prompting text-davinci-002. Under few-shot NL prompting setting, LLama-13B achieves 33.63\%, which is only slightly better than chance (33\%). LLama-70B achieves 43.97\%, around 10\% better than LLaMA-13B and obtains improvements of around 4\% with Chain-of-thought prompting and Tree of Thought prompting. Text-davinci-002 achieves 49.53\% and GPT-3.5 \footnotetext{Hereafter, "GPT-3.5" refers to GPT-3.5-Turbo.} achieves 58.34\%. GPT-4 achieves the best results among GPT series models. 

Incorporating recent prompting strategies increases the performance of vanilla few-shot prompting. Chain-of-thought prompting achieves more than a 4\% increase over GPT-4. Self-consistency (SC) improves chain-of-thought prompting by 0.6\% percent. Tree-of-thought prompting achieves slightly better result than self-consistency with chain-of-thought prompting. For the results of recent methods developed for logical reasoning, LINC \citep{linc} achieves around a 9\% increase over few-shot prompting GPT-4. Both Logic-LM (GPT-4)\citeyearpar{pan-etal-2023-logic} and  
 DetermLR \citeyearpar{sun2023indeterminacy} achieves more than a 13\% increase over few-shot prompting GPT-4, showing the superiority of the methods on logical reasoning. 

%
\nltofol

\paragraph{NL-FOL translation}
Table~\ref{tab:nl-fol-metrics} shows the results of NL-FOL translation.  The syntactic validity scores are around 93\% with both GPT-3.5-Turbo and GPT-4. This indicates that language models with sufficient scales are good at picking up the patterns for FOL formulas and generating syntactically valid FOL formulas. However, GPT-3.5-Turbo and GPT-4 are not yet good at translating an NL story to a logically or semantically similar FOL counterpart, as indicated by the low inference engine execution accuracy score. 



%% file: main/7-analysis.tex
\section{Error Analysis}
Below we provide analysis of our results and key findings, providing additional insights into our dataset \ours and the current capabilities of LLMs in logical reasoning.

\begin{figure}[!t]
    \centering
    {\input{main/figures/analysis-nopremise.tikz}}
    \caption{Accuracies of different models categorized into examples with different reasoning depths.}
    \label{fig:analysis-nopremise}
\end{figure}
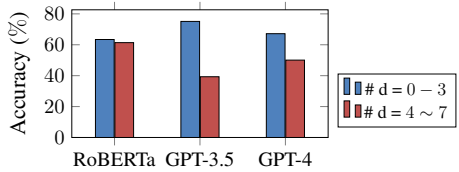

\textbf{Models have higher accuracy on examples with fewer reasoning depths than on those with higher number of reasoing depths}
We show the accuracy categorized by reasoning depths in Figure~\ref{fig:analysis-nopremise}. 
With few-shot prompting, GPT-3.5 and GPT-4 both perform much better on examples with a $0\sim3$ reasoning depth, indicating that examples with a $4\sim7$ reasoning depth pose a challenge to the SoTA LMs. 
With fine-tuning, RoBERTa has slightly higher performance on test examples with $0\sim3$ reasoning depth than on those with $4\sim7$ reasoning depth, but the difference is much smaller.
This indicates that fine-tuning on longer and more difficult reasoning chains in the training set can improve model performance on equally-long test example chains.
We note that the presence and prevalence of these difficult examples are unique to \ours. \ours's unique complexity reveals that current LMs are limited in their ability to extrapolate to longer and more complex reasoning chains, and suggests an avenue for further study.

\wikihyb
\textbf{Models have higher accuracy on WikiLogic than on HybLogic}
As shown in Table~\ref{tab:wikihyb}, in logical reasoning, GPT-3.5 and GPT-4 achieve substantially lower results on HybLogic than on WikiLogic and the result is slightly higher than chance. 
We hypothesize that this is because HybLogic has high logical complexity that the SoTA LLMs like GPT-4 cannot solve yet while WikiLogic examples require shorter reasoning chains which the model is already capable of solving. Moreover, since the examples in WikiLogic are created from scratch by humans, it is possible that LLMs have seen similar texts with similar logical patterns in the training data. 
However, fine-tuning RoBERTa-large achieves higher performance on HybLogic than on WikiLogic. This is likely because HybLogic is created from templates and some of the logical patterns can be learned during fine-tuning.

In NL-FOL translation, performs 10 points better on HybLogic than WikiLogic. This could be because WikiLogic has more distinct and diverse sentence-level logical and language patterns and FOL annotations. WikiLogic has 53 ASTs while HybLogic has 33. Despite being more logically complex on a story level, FOL translations for HybLogic stories have simpler logical structures on a statement level. We include case study for one WikiLogic example and one HybLogic example in Appendix~\ref{app:casestudy} and further analysis on model performance in Appendix~\ref{app:modelperformance}. 

\begin{table}[ht]
\centering
\resizebox{0.4\textwidth}{!}{
\begin{tabular}{l|c}
\toprule
Faulty path & 65\%  \\ 
\midrule
Wrong derivation & 25\%  \\ 
\midrule
Wrong syntactic comprehension & 5\%  \\ 
\midrule
Spurious shortcut & 5\% \\ 
\bottomrule
\end{tabular}
}
\caption{Human evaluation on GPT-4 model outputs with incorrect truth value predictions}
\label{table:human_evaluation_gpt4}
\end{table}

\begin{table}[ht]
\centering
\setlength{\tabcolsep}{3pt}
\small
\begin{tabular}{p{7.5cm}}
\toprule
\textbf{Example} \\ 

Premises
1. All children are human.
2. If someone is underage, then they are a child.
3. People are either underage or of age.
4. If someone is of age, then they can vote.
5. If someone is of age, they can legally get married.
6. If Jack is a child and a human, then Jack is neither able to vote nor able to get married.
Conclusion -> Label:
Jack is able to vote and get married. -> True.\\ 
\midrule
\textbf{GPT-4 Output} \\ 
We know that all children are human (premise 1) and if someone is underage, they are a child (premise 2). 
People are either underage or of age (premise 3).
If someone is of age, they can vote (premise 4) and get married (premise 5). 
If Jack is a child and a human, then Jack is neither able to vote nor get married (premise 6). 
We don't have any information about Jack's age, so we cannot determine if he is a child or of age. 
Therefore, we cannot determine if Jack is able to vote and get married. \\ 
\bottomrule
\end{tabular}
\caption{Case study for the scenario where a model is unable to form the correct reasoning chain. }
\label{tab:case-study}
\end{table}

\textbf{Human evaluation on model outputs}
We conduct human evaluation on the GPT-4 model outputs with wrong truth value predictions. As shown in Table~\ref{table:human_evaluation_gpt4}, approximately 65\% of the time, the model struggles to construct accurate reasoning chains for complex problems with intricate steps, leading to faulty reasoning paths and indicating a limited ability to solve problems with long reasoning chains. In 25\% of cases, erroneous derivations occur within certain reasoning steps, highlighting potential inaccuracies and flaws in logical deductions. 5\% of conclusions in \ours have a complex syntactic structure, posing comprehension challenges for GPT-4. 5\% of outputs show that GPT-4 leverage commonsense reasoning to employ spurious shortcuts that lead to the wrong truth value for the conclusion. We provide a case study for the "Faulty path" scenario in Table~\ref{tab:case-study}. In this instance, the model can perform simple derivations from the premises, like "If someone is of age, they can vote and get married." However, because of the problem's complexity, the model struggles to identify the essential intermediate steps and cannot ascertain the truth value of conclusions, such as "Jack is not a child."


\subsection{Human performance}
We collected truth value annotations of logical reasoning for \ours test set from expert and non-expert annotators. Our expert annotators are computer science college students familiar with FOL. Non-expert annotators are community college or high school students who have not taken the SAT. Both expert and non-expert annotators are native English speakers.
Expert annotations achieve an accuracy of $95.98\%$ while non-expert annotations achieves $61.82\%$, with a gap of {34.16\%}. This shows that sufficient domain knowledge of FOL is necessary for good performance on \ours. The expert and GPT-4 gap is {$31.82\%$}, suggesting significant room for model improvement.

%% file: main/figures/analysis-nopremise.tikz







\begin{tikzpicture}[scale=0.7]
    \begin{axis}[
        width  = 0.4*\textwidth,
        height = 4cm,
        ybar=\pgflinewidth,
        ylabel style={font=\large},
        ylabel = {Accuracy ($\%$)},
        symbolic x coords={RoBERTa, GPT-3.5, GPT-4},
        xtick style={font=\large},
        xtick = data,
        enlarge x limits=0.25,
        ymin=0,
        legend style={
                nodes={scale=0.8, transform shape},
                at={(1.5,0.5)},
                font=\large
        }
    ]
        \addplot[ybar, fill=bblue]
            plot coordinates {(RoBERTa, 63.41) (GPT-3.5,75.12) (GPT-4,67.17)};
        \addplot[ybar, fill=rred]
             plot coordinates {(RoBERTa,61.38) (GPT-3.5,39.31) (GPT-4,50.07)};
        
        \legend{\# d = $0-3$,\# d = $4\sim7$}
    \end{axis}
\end{tikzpicture}

%% file: main/8-conclusion.tex
\section{Conclusion}
We introduced \ours, an expert-written dataset for logical reasoning equipped with FOL formulas. The examples in \ours are created based on real-world knowledge with natural language. It exhibits a large number of distinct logic patterns and a large vocabulary. 
Experiments show that \ours presents a challenge for one of the most capable Large Language Model publicly available. 

\section{Limitations}

We focus on collecting a very high-quality dataset in evaluating logical reasoning rather than merely a large dataset. Optimizing for quality required us to adopt a rigorous annotation process with domain experts selected based on a few important criteria as mentioned in Appendix A: Annotator Selection. Significantly scaling up this process would have required resources beyond our current means and we are unable further expand our dataset for investigating how the size of training data affects the performance of fine-tuning experiments. We encourage the community to apply our annotation protocol to expand this realistic and complex FOL reasoning story set. 

%% file: Appendix.tex
\appendix

\section{Annotator Selection}\label{app:annotator}
Given the complexities of our annotations, we selected annotators based on a few important criteria
1). Our annotators are either college or graduate students who are native English speakers or possess near-native proficiency in English.\footnote{By ``near-native'' we mean with English speaking and understanding ability that closely mirrors that of a native English speakers.} 2). They possess formal education in first-order logic, having either completed relevant coursework or undertaken self-directed studies in first-order logic or semantic parsing. At the NL quality check stage, only annotators who are experts in natural language processing or computational linguistics are involved. For the FOL quality check, only annotators who are experts in first-order logic are involved. We also give the annotators several training sessions on how to write a story, by providing them with detailed annotation guidelines.
All stories and FOL annotations in \ours are written and reviewed by expert annotators, including CS undergraduate and graduate students, and senior researchers, who met the aforementioned criteria. 

\section{HybLogic Template Example}\label{app:template}
An example the resulting template is as follows: 

\begin{small}
\begin{verbatim}
Premises:
All M are P.   All S are M.   
Either S or A.  All A are B. 
All D are B.  No C are B.
a is either a C or a P.

Conclusions:   
[Unknown] a is an S.
[True] If a is either a C or a D, 
then a is not either an A or a B.
\end{verbatim}
\end{small}

\section{Factuality and Bias Elimination Protocol} \label{app:factualityprotocol}
We rewrote those that are not reflective of well-established scientific, historical, or legal facts. We took out stories that had strongly opinionated language and contained gender, racial, and classist biases. We accept certain classes of ``psychologically fundamental generalizations'' \citep{leslie-generics}, however, such as ``Covid is transmitted through the air'' or ``Tigers eat other animals,'' that may not be factually invariant but add logical and semantic nuances to the stories. For stories that pertain to generalization, such as ``All As are Bs,'' we have added specifiers like "all Dan knows" to give a degree of reasonable factuality. For example, ``All science fiction that Dan knows comes from an imaginative process'' has a more reasonable degree of factuality than ``All science fiction comes from an imaginative process.''

\section{Language Quality Control}\label{app:languagequality}

\begin{itemize}[wide, labelwidth=0pt, labelindent=0pt]
    \item We always use ``either-or''  to express exclusive disjunction. We use either ``A or B'' or ``A or B, or both'' to express inclusive disjunction. 
    In English ``or'' itself can be interpreted as either inclusive disjunction or exclusive disjunction. Adding ``or both'' cancels the exclusive disjunction distinctly. However, it is less common in the wild than just using ``or''. we could add ``or both'' if it is important to emphasize the inclusive part semantically or contextually or for factuality; and do not add ``or both'' if it is not.  
We rely on the language model to figure out if it should be inclusive or exclusive, therefore not sacrificing naturalness.
    

    \item It is more natural to say "Some A is B" rather than "there exists an A such that A is B." "All A are B" can be more natural than "If A then B". 
    \item Writing NL sentences that express negation over exclusive-or ("either both or neither") can be cumbersome but we found one natural ways of expressing these situations: "Each morning, John either works out and stretches, or he does neither". 
\end{itemize}
 
Other common issues in NL quality include singular/plural issues, especially in statements that deal with both categories and individual members of those categories; as well as ambiguities resulting from improper introduction of, or failure to introduce, proper nouns.

\section{First-Order Logic}\label{app:firstorderlogic}
\subsection{First-Order Logic VS Natural Language}
 FOL enables deriving facts from other facts \citep{russell-norvig-2010}. In the context of logical reasoning in modern NLP, FOL, as a logical form, is a more explicit logical representation than its NL counterpart and can be used as input to an FOL prover in order to obtain the exact truth values for the conclusions. FOL has no ambiguity while ambiguity can occur at various levels of NLP. FOL can thus be a good interface between how LMs are trained and how logical conclusions are reasoned. 
 
\subsection{FOL definition}
 We include the following operators: negation $\lnot$, conjunction $\land$, disjunction $\lor$, implication $\rightarrow$, universal quantifier $\forall$, existential quantifier $\exists$, equal $=$. Following \cite{russell-norvig-2010}, we consider temporal logic and modal logic as special-purpose logics. Consequently, they are beyond the scope of the definition of first-order logic used in our dataset. 

\subsection{FOL modeling conventions}
We use n-place predicates when applicable for the expressivity of the FOL formulas. However, we do not use the Davidsonian \citep{davidsonsemantics} or neo-Davidsonian semantics \citep{Parsons1990} because translating the majority of the FOL formulas in our dataset only requires one-place and two-place predicates. Therefore the Davidsonian or neo-Davidsonian semantics are not necessary for the expressivity of the FOL formulas. 

For example, "Enjoy dressing up in old-fashioned clothing" is rendered as "Enjoy(x, dressingUp, oldFashionedClothing)".

\section{FOL Annotation Protocol}\label{app:folprotocol}
We therefore design an annotation protocol for first-order logic translation in order to ensure that our FOL translations are as consistent as possible across all examples in our dataset. We highlight a few important strategies used in the annotation protocol. a). First-order logic formulas need to preserve as much as possible the semantics of natural language sentences.  b). First-order logic formulas should stay as faithful to the structure of the
original NL sentence as possible. c). Semantic decomposition is not needed unless necessary for maintaining the NL expressivity. This means that "John is a bachelor" can be translated into FOL simply as "\texttt{Bachelor(John)}". d). In terms of abstraction, we neglect tense and remove all the plural forms of verbs. 

\section{FOL Inference Engine}\label{app:inferencengine}
Although there are many provers widely used in the community ~\cite{prover9-mace4, tptp, nipkow2002isabelle} , we adopt the inference engine provided in the Stanford CS221 course page\footnote{\url{https://stanford-cs221.github.io/spring2022/assignments/logic/index.html}}, which is a compact module designed specifically for procesing first-order logic statements. 
The inference engine does not support input in the FOL syntax adopted by standard education material \cite{russell-norvig-2010}, which is used in our dataset. We therefore developed a FOL parser in order to convert the FOL formulas written by humans to the input format of the inference engine. 
The converter is a semantic parser tool written in Python. Although LLMs such as GPT-4 can be utilized to conduct the conversion, it is hard to ensure the GPT-4 outputs are always correct. 

Proving a story requires three steps. First, the FOL statements of the premises and conclusions of a story annotated by humans are converted to Python code. Then, the code snippets are used as input to the theorem prover. 
Finally, the theorem prover outputs whether the conclusions are True / False / Unknown, based on the premises.

\section{Distribution of Readability}\label{app:readability}
We show the distribution of readability in Figure~\ref{fig:complexity}. 
\dalechall

\section{Case study}\label{app:casestudy}

\casestudy

\casestudytwo
Table~\ref{tab:case_study1} shows a story from WikiLogic along with the GPT-4 and RoBERTa-Large predictions. Conclusion A is True given premises 5 and 3. From the premises, it cannot be determined if Cerura vinula has thin antennae or if it is a pest. Thus conclusions B and C are Unknown. GPT-4 predictions are correct for conclusions A and C while RoBERTa predictions are wrong for all conclusions. 

Table~\ref{tab:case_study2} shows a story from HybLogic with a more complex FOL reasoning process. Inferred from premises 4 and 5, James does not perform better than others. With premises 3, 2 and 1, we know that James is not good at time management. Therefore, conclusion B is False. It cannot be determined if James exercises every week, thus the first conclusion is Unknown. The truth value of $p \rightarrow q$ is the same as $\lnot p \lor q$. It is not true that James does not perform better than others. It is also false that James exercises every week and is good at time management. Thus conclusion C is False. For this example, GPT-4 predicted the correct truth value only for conclusion A and RoBERTa made correct predictions for conclusions A and B.

\section{Model Performance Analysis}\label{app:modelperformance}
\paragraph{Models have more tendency to predict ``True'' compared with ``False'' or ``Unknown'' labels}
Confusion matrices in Figure~\ref{fig:cf} for the fine-tuning and 8-shot NL prompt results both show that LLMs are significantly better at making the correct predictions for conclusions with labels of True than the conclusions with labels of False or Unknown. The accuracy on examples with False or Unknown conclusions is 61.9\% with fine-tuning and 54.0\% with few-shot prompting.
They also tend to make more predictions of True than the other labels.
\cf

\paragraph{Model performance is not affected by the premise ordering}
To test if the premise ordering in \ours has spurious correlations with the conclusion label which a model can exploit, we shuffle the input premises to evaluate models.
We find that accuracy increases or decreases by roughly 1\% in most settings compared to our unshuffled premises.
This indicates that the ordering of premises in \ours examples does not yield significant information about the label, and thus models will not be able to use the premise ordering as a strong heuristic or statistical feature for its predictions.

\nlandfolprompt   
\paragraph{Using both NL sentences and FOL formulas in the prompt performs better}
FOL formulas have a clearer and more straightforward logical structure than NL sentences. Therefore, we test GPT-3.5 and GPT-4 with another two settings for truth value prediction using few-shot prompting: 1) using only FOL formulas in the prompt; 2) using both NL sentences and FOL formulas by concatenating each NL sentence and its annotated FOL statement. As shown in Table~\ref{tab:nl+fol-prompt}, the performance slightly increases in the NL+FOL setting for GPT-4 while GPT-3.5 performs worse in both the NL+FOL and the FOL-only settings. In other words, FOL always serves as additional useful information for GPT-4, but not for GPT-3.5 regardless of whether FOL is concatenated with NL. 
This observation resonates with the finding that GPT-4 performs much better than GPT-3.5 on code-related tasks \citep{ni2023l2ceval}.